\documentclass[11pt]{article}

\usepackage[preprint]{acl}

\usepackage{times}
\usepackage{latexsym}

\usepackage[T1]{fontenc}

\usepackage[utf8]{inputenc}

\usepackage{microtype}

\usepackage{inconsolata}

\usepackage{graphicx}
\usepackage[utf8]{inputenc} 
\usepackage[T1]{fontenc}    
\usepackage{hyperref}       
\usepackage{url}            
\usepackage{booktabs}       
\usepackage{amsfonts}       
\usepackage{nicefrac}       
\usepackage{microtype}      
\usepackage{xcolor}         
\usepackage{marvosym}
\usepackage{multirow}
\usepackage{makecell}
\usepackage{colortbl}
\usepackage{tcolorbox}
\usepackage{longtable}
\usepackage{subcaption}
\usepackage{pifont}
\usepackage{framed}
\usepackage[utf8]{inputenc}
\usepackage{amsmath}
\usepackage{enumitem}
\usepackage{color}

\usepackage{tikz}
\usepackage[edges]{forest}
\usepackage{pgfplots}
\usepackage{bm}
\usepackage{scalefnt}
\usepackage[framemethod=tikz]{mdframed}
\usepackage{subcaption}
\definecolor{lgreen}{rgb}{0.89,0.94,0.85}
\definecolor{lred}{rgb}{0.98, 0.90, 0.84}
\definecolor{lyellow}{rgb}{1.00, 0.95, 0.80}
\definecolor{lblue}{rgb}{0.85, 0.89, 0.95}
\definecolor{hidden-draw}{RGB}{20,68,106}
\definecolor{hidden-pink}{RGB}{255,245,247}

\title{The Imperative of Conversation Analysis in the Era of LLMs:\\ A Survey of Tasks, Techniques, and Trends}

\author{%
  Xinghua Zhang, 
  Haiyang Yu,
  Yongbin Li\thanks{Corresponding author.},
  Minzheng Wang,
  Longze Chen,
  Fei Huang\\
  Alibaba Group, China\\
  \small{
\Letter: \{\texttt{zhangxinghua.zxh}, \texttt{yifei.yhy}, \texttt{shuide.lyb}, \texttt{wangminzheng.wmz}, \texttt{chenlongze.clz}, \texttt{f.huang}\}@alibaba-inc.com}
}

\begin{document}
\maketitle

\begin{abstract}
In the era of large language models (LLMs), a vast amount of conversation logs will be accumulated thanks to the rapid development trend of language UI. Conversation Analysis (CA) strives to uncover and analyze critical information from conversation data, streamlining manual processes and supporting business insights and decision-making. The need for CA to extract actionable insights and drive empowerment is becoming increasingly prominent and attracting widespread attention.
However, the lack of a clear scope for CA leads to a dispersion of various techniques, making it difficult to form a systematic technical synergy to empower business applications.
In this paper, we perform a thorough review and systematize CA task to summarize the existing related work. Specifically, we formally define CA task to confront the fragmented and chaotic landscape in this field, and derive four key steps of CA from conversation scene reconstruction, to in-depth attribution analysis, and then to performing targeted training, finally generating conversations based on the targeted training for achieving the specific goals.
In addition, we showcase the relevant benchmarks, discuss potential challenges and point out future directions in both industry and academia.
In view of current advancements, it is evident that the majority of efforts are still concentrated on the analysis of shallow conversation elements, which presents a considerable gap between the research and business, and with the assist of LLMs, recent work has shown a trend towards research on causality and strategic tasks which are sophisticated and high-level.
The analyzed experiences and insights will inevitably have broader application value in  business operations that target conversation logs.
\end{abstract}
\section{Introduction}
With the development of large language models (LLMs), the next generation of system interactions is rapidly advancing towards natural language conversation interfaces (Language UI)~\cite{wang2023enabling,dong2023towards,wu2024autogen}, leading to a vast amount of natural language interaction logs. It is more valuable to extract, summarize, analyze, and reason from these conversation\footnote{In the literature, ``conversation'' and ``dialogue'' are both commonly used, referring to different communication types. CA aims to analyze any form and type of communications, such as human-human, human-machine, multi-party~\cite{ganesh2023survey}, etc. We refer to them collectively as ``conversation'' in this paper.} logs, which produces numerous new applications such as system optimization~\cite{xu2020end,liu2023one}, customer operations~\cite{wang2020sentiment,zou2021topic}, and demand insights~\cite{deriu2021survey,li2021seamlessly,he2023conversation}.
The goal of conversation analysis (CA) is to mine and analyze out critical information (such as customer profiles, purchase intentions, mood changes, sales skills, shortcomings and corresponding improvement proposals, etc.) from conversational data, which contributes to simplifying manual processes and assisting in business insights and decision-making.

\begin{figure}[t]
\centering
\includegraphics[width=0.482\textwidth]{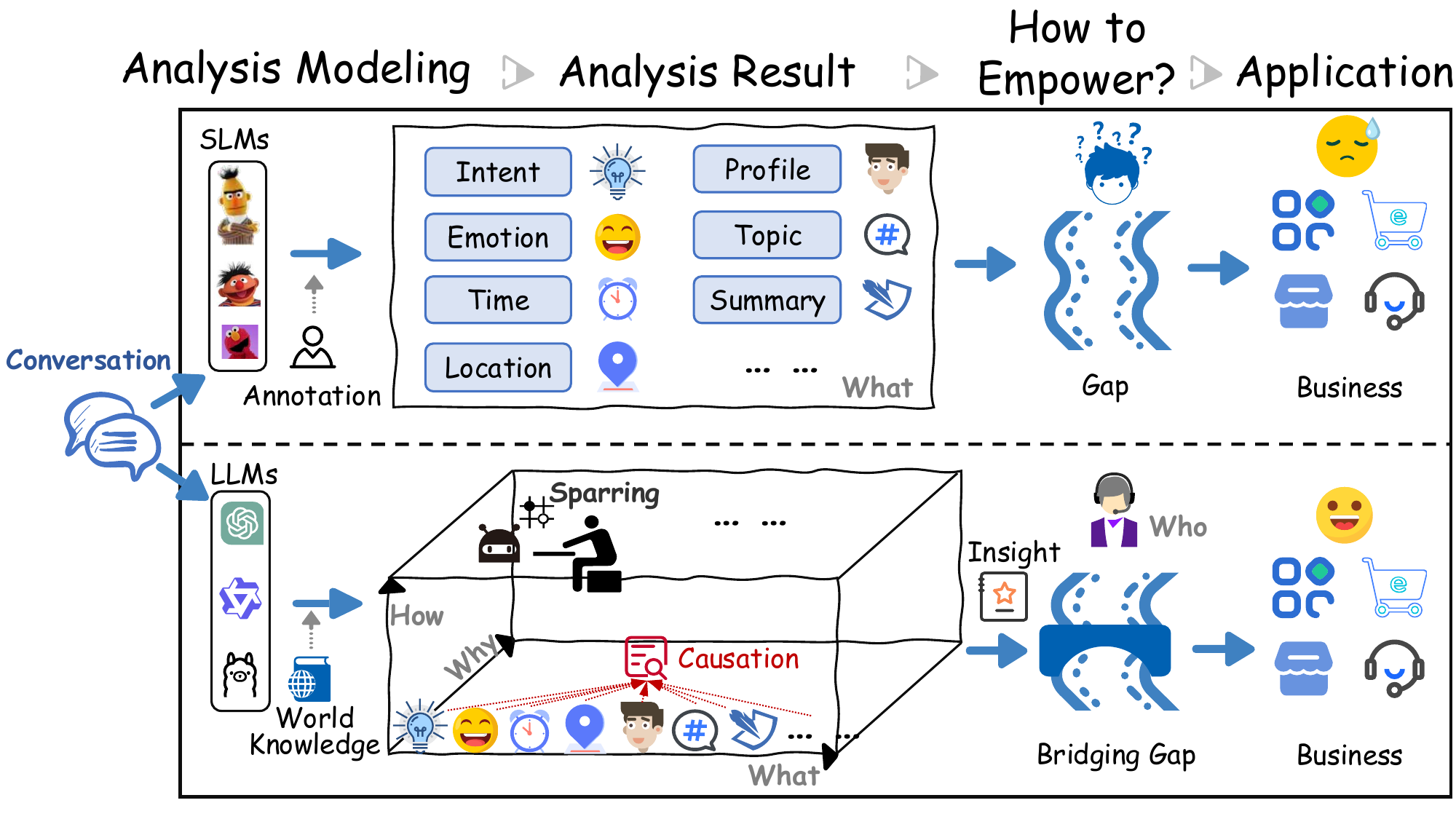} 
\caption{Conversation analysis results are usually shallow in the era of small language models (SLMs) and cannot well empower the real-world applications, while large language models (LLMs) possess world knowledge, and can contribute to in-depth attributions and provide constructive insights or experiences.}
\label{ca_intro}
\end{figure}

In the era of small language models (SLMs), conversational data based analysis did not form a systematic task but was broken down into atomic tasks with strict output constraints, such as intent classification, slot filling~\cite{qin-etal-2020-agif,louvan2020recent,jiang-etal-2023-spm}, and conversation summarization~\cite{fabbri2021convosumm,ouyang-etal-2023-compositional,ramprasad-etal-2024-analyzing} as shown in the upper part of Figure~\ref{ca_intro}. Meanwhile, the analysis results generated by these atomic tasks are generally shallow, flat and there is a significant gap between the results and the genuine needs of the business, failing to address the industry's pain points. The core challenge was that small models could not model the world, thus they were unable to reconstruct the actual scene, such as conversation participants, scenarios, insightful rules from the ``conversation''. However, as shown in the lower part of Figure~\ref{ca_intro}, with the rise of LLMs which fulfil a wealth of world knowledge~\cite{zhang2023large,yildirim2024task,zhao2024expel}, it has become possible to infer detailed ``real-world scene'' from conversations and generate multifaceted and in-depth conversation analysis results, from ``{\it What is the scene elements behind the conversation}'', to ``{\it Why do these elements exist}'', then to ``{\it How to deal with experience or problem}'', finally answering ``{\it Who should participant become}''. The more detailed and accurate the inferred conversation information is, the more precise and valuable insights we can provide for business applications.

Despite achieving significant progress in the era of LLMs, CA continues to encounter certain difficulties: 
(1) \textbf{Definition}: there is a lack of a systematic definition of CA, which leads to a dispersion in the optimization objectives and technical aspects.
(2) \textbf{Data}: there is a scarcity of CA datasets that encompass all essential elements of conversations, making it challenging to accurately model and evaluate the conversation background information, which affects the development of CA.
(3) \textbf{Method}: Unlike flat text or documents, conversations inherently possess characteristics such as multi-turn interactions, strict contextual dependencies, implicit ambiguity, and colloquial language, which necessitate deeper modeling techniques.
(4) \textbf{Application}: the current analysis yields rather shallow results when it comes to gauging emotions, opinions, and intentions, lacking a unified, profound, and constructive analytical perspective, which leads to the extensive gap between the realm of research and practical application.

However, there is currently a lack of technical surveys to showcase the research progress and to outline the research blueprint for CA. To bridge this gap, we make the first attempt to give the thorough review of existing research related to CA.
According to the procedure of conversation analysis, the field of CA is segmented into four essential parts: (1) {\it Scene Reconstruction} for inferring the latent scene elements in conversation and relatively shallow analyses, (2) {\it Causality Analysis} for undertaking in-depth attributions, and (3) {\it Skill Enhancement} for training employees or models based on the attributions, and (4) {\it Conversation Generation} following the improved skills for generating conversations.
Centered around the \textbf{goals} of system optimization (such as increasing customer retention rate), these four procedures seamlessly integrate from preliminary reconstruction, to in-depth attribution, and then to the targeted capability enhancement, ultimately re-generating the conversations and are continuously iterated in pursuit of more effectively achieving the goals.
The major contributions of this survey are summarized as follows:
\begin{itemize}[leftmargin=*]

\item \textbf{First technical review}: To our knowledge, we present the first technical CA survey from the viewpoint of goal-directed analysis with four critical procedures, integrating related techniques.

\item \textbf{Systematic task definition}: Centered around the goal improvements, we give the clear task definition and optimization objectives of CA.

\item \textbf{Abundant data resource}: We organize the existing data resources and evaluation criteria.

\item \textbf{Insightful discussions}: We discuss the existing challenges and shed light on the future directions of CA in the era of LLMs.

\end{itemize}
\begin{figure*}[t]
\centering
\includegraphics[width=0.9\textwidth]{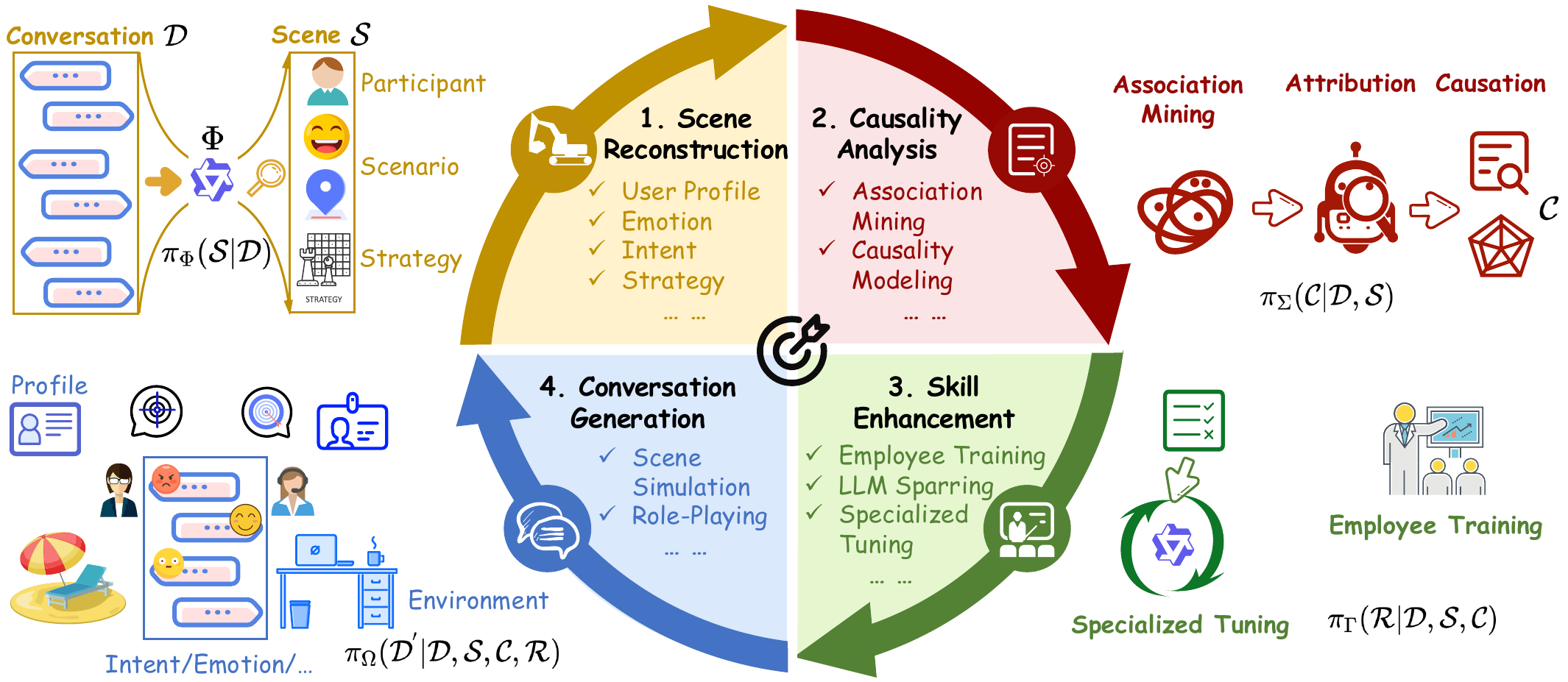} 
\caption{Overview of Conversation Analysis, which consists of 1) Scene Reconstruction, 2) Causality Analysis, 3) Skill Enhancement, and 4) Conversation Generation, centered around the goal-directed optimization.}
\label{ca_diagram}
\end{figure*}

\section{Concept and Formulation}

This section gives the definition of conversation analysis (CA), and then formally describes the critical procedures within CA.

\textbf{Conversation analysis aims to identify critical information from human-human, human-machine, machine-machine, and multi-party conversations, derive the underlying causes, and develop the solutions to drive relevant improvements for more effective goal achievement continuously}, such as {\it elevating customer experience}, {\it reducing complaint rate}. As shown in Figure~\ref{ca_diagram}, CA typically includes several components/procedures: 1) {\it Scene Reconstruction}, 2) {\it Causality Analysis}, 3) {\it Skill Enhancement}, and 4) {\it Conversation Generation}.
Specifically, CA system first analyzes the latent scene elements $\mathcal{S}$ such as participant profile, emotion, intent within the conversation, and then the factors behind these elements are uncovered systematically, drafting a comprehensive analytical report or making corresponding attributions $\mathcal{C}$. Furthermore, built upon these factors, the latent insights $\mathcal{R}$ of system can be improved via intelligent agent or employee training to facilitate the attainment of our goals $\mathcal{G}$. Finally, CA system generates the conversational content $\mathcal{D}$ following the improved insights for next-round iteration.

\textbf{Scene Reconstruction.} This procedure tends to attain the scene elements $\mathcal{S}$ (e.g., participants, scenarios including emotion, intent, environment, etc) from the conversational content $\mathcal{D}$. In fact, $\mathcal{D}$ arises within the specific scene $\mathcal{S}$, and this process aims to reconstruct the actual $\mathcal{S}$ according to $\mathcal{D}$ through tailor-designed modeling $\Phi$. The scene reconstruction procedure can be formalized as $\pi_{\Phi}(\mathcal{S}|\mathcal{D})$.
This procedure takes conversational data $\mathcal{D}$ as input and then restores the objective elements within it for contributing to insightful analyses in the subsequent steps.

\textbf{Causality Analysis.} This procedure seeks to delve deeply into the rationales $\mathcal{C}$ behind the components $\mathcal{S}$, such as the reasons for participants' attitude shifts, persuasive strategies. To achieve this, some causal modeling methods $\Sigma$ can be developed. For instance, association mining can be performed within multi-group conversations and their corresponding scene components, and insightful reports and causation $\mathcal{C}$ are reached by condensing these items. The causality analysis procedure can be formalized as $\pi_{\Sigma}(\mathcal{C}|\mathcal{D},\mathcal{S})$.
Causality analysis takes conversational content $\mathcal{D}$ and $\mathcal{S}$ reconstructed in \S {\it Scene Reconstruction} as input and then derives more precise factors and induced reasons which contribute to the scene elements $\mathcal{S}$.

\textbf{Skill Enhancement.} This procedure strives to optimize the entire system towards the goal according to the causation $\mathcal{C}$. Concretely, for human participants such as call-center employees, there is a need for relevant departments to provide training $\Gamma$ for human participants based on the feedback. As for AI agent, we need experts to optimize the agent model $\Gamma$ for better insights $\mathcal{R}$ towards achieving the goals according to $\mathcal{C}$. The skill enhancement procedure can be formalized as $\pi_{\Gamma}(\mathcal{R}|\mathcal{D},\mathcal{S},\mathcal{C})$.
Skill enhancement refers $\mathcal{D}$, $\mathcal{S}$, and $\mathcal{C}$ analyzed in \S {\it Causality Analysis} to train the personnel or optimize the model for attaining the refined $\mathcal{R}$, which guides the direction towards achieving the goal.

\textbf{Conversation Generation.} This procedure aims to collect the conversational content $\mathcal{D}$ from real-world conversation data after employee training, such as online customer service and hotline data.
In addition, we also drive the models to synthesize data based on the refined insights $\mathcal{R}$ (including scene elements, fundamental abilities, etc.) within the conversation via the techniques such as role-playing model $\Omega$. The conversation generation procedure can be formalized as $\pi_{\Omega}(\mathcal{D}^{'}|\mathcal{D},\mathcal{S},\mathcal{C},\mathcal{R})$.
The conversation generation results in the collection of vast amounts of data, supporting thorough analysis of conversation and serving as the most direct evidence of goal achievement.

According to multi-action reinforcement learning~\cite{LEI2022117796,lwenhao2022-structured,wang2022man}, the entire process of conversation analysis can be expressed formally as an Markov decision process (MDP) defined by a tuple ($\mathcal{D}$, $\mathcal{A}$, $P$, $r$). $\mathcal{D}$ is the state and represents the conversational content in CA. The action set $\mathcal{A}=\mathcal{S}\times\mathcal{C}\times\mathcal{R}$ is three-dimensional, in which $\mathcal{S}$ is the scene reconstruction sub-action set, $\mathcal{C}$ is the causality analysis sub-action set, and $\mathcal{R}$ is the skill enhancement sub-action set. $P$ is the transition probability distribution and $r$ is the reward function.
The MDP with a multi-action space is referred to as multiple MDPs (MMDPs) in ~\citet{LEI2022117796} and we defines the MMDPs over the states, actions, and reward in CA as follows:
\begin{itemize}
    \item \textbf{States:} The global state is $\mathcal{D}_t$ at timestep $t$ which denotes the conversational content generated subsequent to the $t$-th iteration. Following ~\citet{wang2022man}, the local states corresponding to the sub-action sets are enriched by a concatenation of the state $\mathcal{D}_t$ and the history of previous $k$-$1$ dimension sub-action choice.
    \item \textbf{Actions:}  The actions at timestep $t$ consist of scene reconstruction action $\mathcal{S}_t$, causality analysis action $\mathcal{C}_t$, and skill enhancement action $\mathcal{R}_t$, which correspond to analyzing the conversation scene, latent causation, and enhanced insights separately.
    \item \textbf{Policy:} The policy $\pi_{\Theta}(\mathcal{A}|\mathcal{D})$ composes of three stochastic sub-policies $\pi_{\Phi}(\mathcal{S}|\mathcal{D})$, $\pi_{\Sigma}(\mathcal{C}|\mathcal{D},\mathcal{S})$, and $\pi_{\Gamma}(\mathcal{R}|\mathcal{D},\mathcal{S},\mathcal{C})$, where $\Phi$, $\Sigma$, and $\Gamma$ are respectively models for scene reconstruction, causality analysis, and skill enhancement. $\Theta = [\Phi, \Sigma, \Gamma]$ is the entire set of all models. $\pi_{\Theta}(\mathcal{A}|\mathcal{D})$ selects a series of actions $\mathcal{A}$ ($\mathcal{S}$, $\mathcal{C}$, $\mathcal{R}$) in their respective sub-action space.
    \item \textbf{Transition:} At timestep $t$, given the sub actions $\mathcal{S}_t$, $\mathcal{C}_t$, $\mathcal{R}_t$, and state $\mathcal{D}_t$, the global state is updated as $\mathcal{D}_{t+1}$. The process happens to be the conversation generation $\pi_{\Omega}(\mathcal{D}^{'}|\mathcal{D},\mathcal{S},\mathcal{C},\mathcal{R})$, which can be reformulated as $P(\mathcal{D}_{t+1}|\mathcal{D}_t,\mathcal{S}_t,\mathcal{C}_t,\mathcal{R}_t)$.
    \item \textbf{Reward:} CA is to learn a whole policy to make the conversations achieve the goals such as {\it elevating customer experience}, {\it reducing complaint rate}, and its reward can be formalized as $\sum_{t} r(\mathcal{D}_t, \mathcal{S}_t,\mathcal{C}_t,\mathcal{R}_t)$.
\end{itemize}

\tikzstyle{my-box}=[
    rectangle,
    draw=hidden-draw,
    rounded corners,
    text opacity=1,
    minimum height=1.5em,
    minimum width=5em,
    inner sep=2pt,
    align=center,
    fill opacity=.5,
    line width=0.8pt,
]
\tikzstyle{leaf}=[my-box, minimum height=1.5em,
    fill=hidden-pink!80, text=black, align=center,font=\normalsize,
    inner xsep=2pt,
    inner ysep=4pt,
    line width=0.8pt,
]
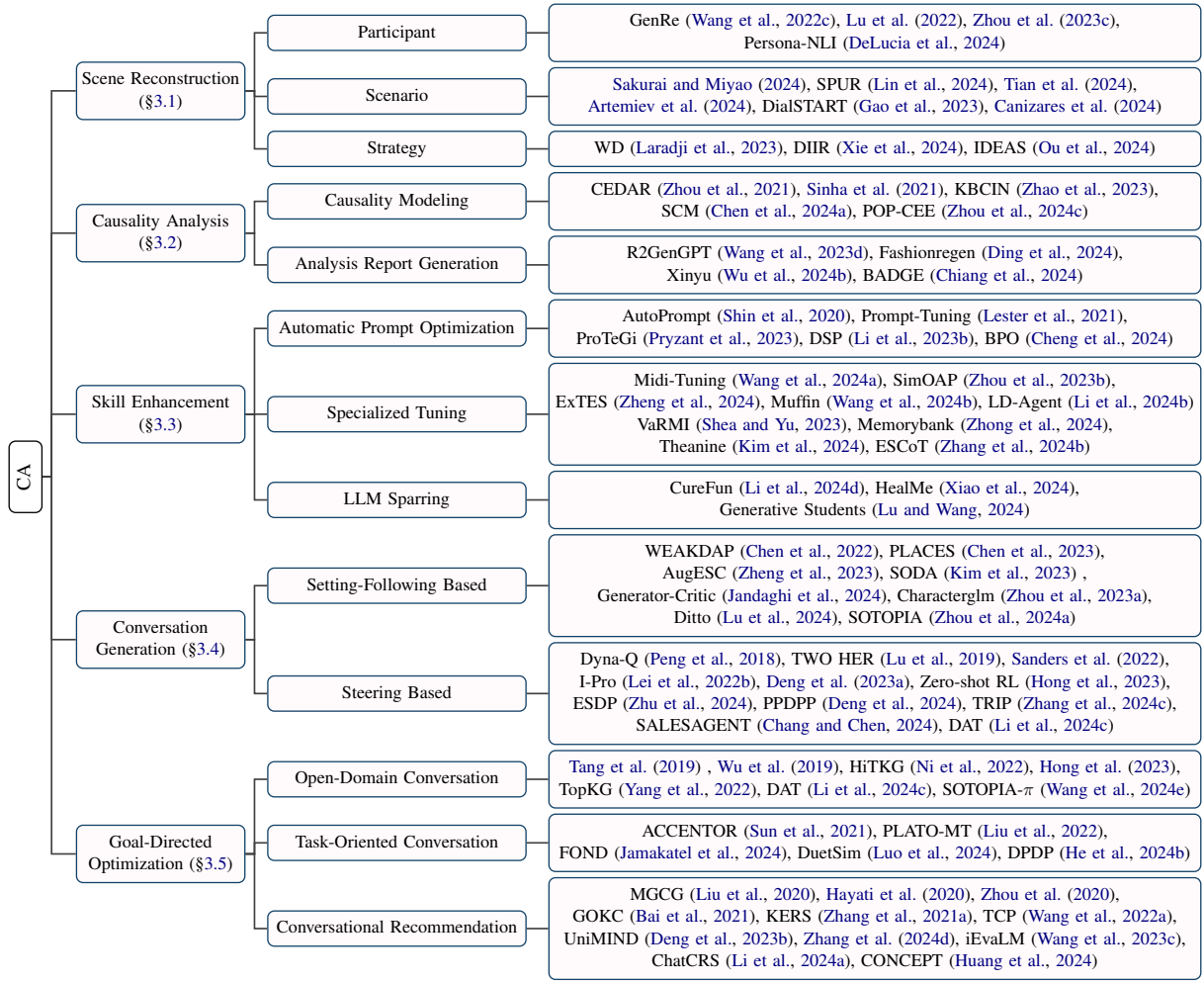
\begin{figure*}[t!]
    \centering
    \resizebox{1.0\textwidth}{!}{
        \begin{forest}
            forked edges,
            for tree={
                grow'=0,
                draw,
                reversed=true,
                anchor=base west,
                parent anchor=east,
                child anchor=west,
                base=center,
                font=\large,
                rectangle,
                rounded corners,
                align=center,
                minimum width=4em,
                edge+={darkgray, line width=1pt},
                s sep=3pt,
                inner xsep=2pt,
                inner ysep=3pt,
                line width=0.8pt,
                ver/.style={rotate=90, child anchor=north, parent anchor=south, anchor=center},
            },
            where level=1{leaf,text width=9em,font=\normalsize,}{},
            where level=2{leaf,text width=14em,font=\normalsize,}{},
            where level=3{leaf,text width=8em,font=\normalsize,}{},
            where level=4{leaf,text width=5em,font=\normalsize,}{},
			[
			    CA, ver
			    [
			        Scene Reconstruction\\ (\S \ref{tca:sr})
			        [
			            Participant
			            [
                          GenRe~\cite{wang-etal-2022-extracting}{, }\citet{lu2022partner}{, }\citet{zhou2023learning}{, }\\Persona-NLI~\cite{delucia2024using}
                            , leaf, text width=35.9em
			            ]
			        ]
			        [
			            Scenario
			            [
			                \citet{sakurai-miyao-2024-evaluating}{, }SPUR~\cite{lin-etal-2024-interpretable}{, }\citet{tian-etal-2024-dialogue}{, }\\ \citet{artemiev-etal-2024-leveraging}{, }DialSTART~\cite{gao2023udts}{,} \citet{canizares2024measuring}
                            , leaf, text width=35.9em
			            ]
			        ]
			        [
			            Strategy
			            [
			                WD~\cite{laradji2023workflow}{, }DIIR~\cite{xie-etal-2024-shot-dialogue}{, }IDEAS~\cite{ou2024inductivedeductivestrategyreusemultiturn}
                            , leaf, text width=35.9em
			            ]
			        ]
			    ]
                [
			        Causality Analysis\\ (\S \ref{tca:ca})
			        [
			            Causality Modeling
			            [
			                CEDAR~\cite{zhou-etal-2021-probing-commonsense}{, }\citet{sinha2021explaining}{, }KBCIN~\cite{zhao2023knowledge}{, }\\SCM~\cite{chen2024learning}{, }POP-CEE~\cite{zhou-etal-2024-pop}
                            , leaf, text width=35.9em
			            ]
			        ]
			        [
			            Analysis Report Generation
			            [
			                 R2GenGPT~\cite{wang2023r2gengpt}{, }Fashionregen~\cite{ding2024fashionregen}{, }\\Xinyu~\cite{wu2024xinyu}{, }BADGE~\cite{chiang2024badge}
                            , leaf, text width=35.9em
			            ]
			        ]
			    ]
                [
			        Skill Enhancement\\ (\S \ref{tca:se})
			        [
			            Automatic Prompt Optimization
			            [
			                AutoPrompt~\cite{shin-etal-2020-autoprompt}{, }Prompt-Tuning~\cite{lester-etal-2021-power}{, }\\ProTeGi~\cite{pryzant-etal-2023-automatic}{, }DSP~\cite{NEURIPS2023_c5601d99}{, }BPO~\cite{cheng-etal-2024-black}
                            , leaf, text width=35.9em
			            ]
			        ]
			        [
			            Specialized Tuning
			            [
			                Midi-Tuning~\cite{wang-etal-2024-instruct}{, }SimOAP~\cite{zhou-etal-2023-simoap}{, }\\ExTES~\cite{zheng-etal-2024-self}{, }Muffin~\cite{wang-etal-2024-muffin}{, }LD-Agent~\cite{li2024helloagain}\\VaRMI~\cite{shea-yu-2023-building}{, }Memorybank~\cite{Zhong_Guo_Gao_Ye_Wang_2024}{, }\\Theanine~\cite{kim2024theanine}{, }ESCoT~\cite{zhang-etal-2024-escot}
                            , leaf, text width=35.9em
			            ]
			        ]
			        [
			            LLM Sparring
			            [
			                CureFun~\cite{li2024leveraging}{, }HealMe~\cite{xiao-etal-2024-healme}{, }\\ Generative Students~\cite{Lu_2024}
                            , leaf, text width=35.9em
			            ]
			        ]
			    ]
                [
			        Conversation\\ Generation (\S \ref{tca:cg})
			        [
			           Setting-Following Based
			            [   WEAKDAP~\cite{chen2022weakly}{, }PLACES~\cite{chen-etal-2023-places}{, }\\AugESC~\cite{zheng-etal-2023-augesc}{, }SODA~\cite{kim-etal-2023-soda}
                                {, }\\Generator-Critic~\cite{jandaghi-etal-2024-faithful-persona}{, }Characterglm~\cite{zhou2023characterglm}{, }\\ Ditto~\cite{lu2024large}{, }SOTOPIA~\cite{zhousotopia}
                            , leaf, text width=35.9em
			            ]
			        ]
                    [
			            Steering Based
			            [
			                Dyna-Q~\cite{peng-etal-2018-deep}{, }TWO HER~\cite{lu2019goal}{, }\citet{sanders-etal-2022-towards}{, }\\I-Pro~\cite{lei2022interacting}{, }\citet{deng-etal-2023-prompting}{, }Zero-shot RL~\cite{hong2023zeroshot}{, }\\ESDP~\cite{zhu2024emotion}{, }PPDPP~\cite{deng2024plugandplay}{, }TRIP~\cite{zhang2024strength}{, }\\SALESAGENT~\cite{chang-chen-2024-injecting}{, }DAT~\cite{li2024dialogue}
                            , leaf, text width=35.9em
			            ]
			        ]
			    ]
                [
			        Goal-Directed\\ Optimization (\S \ref{tca:gdo})
			        [
			            Open-Domain Conversation
			            [
			                \citet{tang2019target} {, }\citet{wu2019proactive}{, }HiTKG~\cite{Ni_Pandelea_Young_Zhou_Cambria_2022}{, }\citet{hong2023zeroshot}{, }\\TopKG~\cite{yang2022topkg}{, }DAT~\cite{li2024dialogue}{, }SOTOPIA-$\pi$~\cite{wang-etal-2024-sotopia}
                            , leaf, text width=35.9em
			            ]
			        ]
			        [
			            Task-Oriented Conversation
			            [
			                ACCENTOR~\cite{Sun_Moon_Crook_Roller_Silvert_Liu_Wang_Liu_Cho_Cardie_2021}{, }PLATO-MT~\cite{liu2022go}{, }\\FOND~\cite{ijcai2024p870}{, }DuetSim~\cite{luo2024duetsim}{, }DPDP~\cite{he-etal-2024-planning}
                            , leaf, text width=35.9em
			            ]
			        ]
			        [
			            Conversational Recommendation
			            [
			                MGCG~\cite{liu2020towards}{, }\citet{hayati2020inspired}{, }\citet{zhou2020towards}{, }\\GOKC~\cite{bai2021learning}{, }KERS~\cite{zhang2021kers}{, }TCP~\cite{wang2022follow}{, }\\UniMIND~\cite{deng2023unified}{, }\citet{zhang2024goal}{, }iEvaLM~\cite{wang2023rethinking}{, }\\ChatCRS~\cite{li2024incorporating}{, }CONCEPT~\cite{huang2024concept}
                            , leaf, text width=35.9em
			            ]
			        ]
			    ]
			]
            \end{forest}
    }
    \caption{Taxonomy for Conversation Analysis (CA).}
    \label{fig:taxonomy}
\end{figure*}

\section{Taxonomy of Conversation Analysis}
This section mainly describes the research related to CA according to the taxonomy of analysis procedure, including scene reconstruction (\S \ref{tca:sr}), causality analysis (\S \ref{tca:ca}), skill enhancement (\S \ref{tca:se}), and conversation generation (\S \ref{tca:cg}) for goal-directed optimization (\S \ref{tca:gdo}).

\subsection{Scene Reconstruction}
\label{tca:sr}
A conversation contains conversational content, participants, scenarios (such as location, topic, atmosphere, emotion, intent) and so on. The elements other than the conversational content are collectively referred to as the scene. Scene reconstruction aims to mine these scene elements which are the background information of a conversation.

\textbf{Participant.}
\citet{tigunova2020extracting} investigated what salient attributes can be extracted from conversational texts and which tools may be applicable for that.
\citet{wang-etal-2022-extracting} introduced the tasks of extracting and inferring personal attributes from conversations, such as hobbies, pets, family, likes and dislikes. They also utilized constrained attribute generation based on autoregressive language model with the discriminative reranker.
\citet{lu2022partner} used automatic partner personas generation to enhance the response generation.
\citet{zhou2023learning} proposed a method which learnt to predict persona information based on the conversation history.
\citet{delucia2024using} post-hoc adapted a trained persona extraction model to a new domain by a natural language inference method.

\textbf{Scenario.}
\citet{sakurai-miyao-2024-evaluating} modified the existing datasets of persuasive conversation and created datasets using a multiple-choice paradigm to evaluate LLMs' intention detection capability in conversation.
\citet{lin-etal-2024-interpretable} introduced SPUR to estimate user satisfaction, which developed an iterative prompting framework that uses an LLM to firstly extract signals of satisfaction in a labeled training set, then summarize the reasons into rubrics, and lastly apply the rubrics to predict satisfaction labels.
\citet{tian-etal-2024-dialogue} proposed an LLM-based approach with role-oriented routing and fusion generation to utilize mixture of experts for conversation summarization.
Due to the variety in segmentation granularity, as well as the lack of diverse annotated corpora, there is no universal models can be easily applied to domain-specific applications. \citet{liu-etal-2023-joint} firstly introduced a joint model for conversation segmentation and topic classification, conducting a case study on healthcare follow-up calls for diabetes management.
DialSTART~\citep{gao2023udts} learned topic-aware utterance representations from unlabeled conversation data through adjacent utterance matching and pseudo-segmentation. These topic-aware utterance representations are then combined with the conversation coherence to perform unsupervised segmentation.
\citet{artemiev-etal-2024-leveraging} proposed a novel conversation summarization-based approach, focusing on transcribed spoken conversations, achieving significant improvements in unsupervised topic segmentation.
\citet{feder2022causal} adopted the clustering method to associate toxic and non-toxic groupings among the formerly suspended users, tackling challenges for causal assumption.
\citet{canizares2024measuring} developed the clustering methods to help in grouping chatbots along their conversation topics and design features.

\textbf{Strategy.}
\citet{laradji2023workflow} introduced a new task called the workflow discovery which aims to discover the set of actions that have been taken to solve a specific requirement.
\citet{xie-etal-2024-shot-dialogue} proposed DIIR framework to learn and apply conversation strategies in the form of natural language inductive rules from expert demonstrations, which generates and verifies a strategy description given a context and its corresponding gold response.
\citet{ou2024inductivedeductivestrategyreusemultiturn} derived instructional strategies from diverse real instruction conversations, which were then deductively exploited to new conversation scenarios, contributing to high-quality instructions.


\subsection{Causality Analysis}
\label{tca:ca}
The process of conversation interaction is  sophisticated with the latent induced factors. Analyzing these factors has important guiding significance for gaining a deeper understanding of the causes and development during conversations.
By performing causality modeling, and  generating analysis reports, causality analysis aims to delve into the causes which lead to the emergence of scene elements, such as emotional changes and user personality.


\textbf{Causality Modeling.}
\citet{zhou-etal-2021-probing-commonsense} explored why models respond as they do by probing their understanding of commonsense reasoning that elicits proper responses. The authors framed commonsense as a latent variable to formally define the problem which aimed to generate commonsense explanations for responses.
\citet{sinha2021explaining} analyzed why a conversation ends with a particular sentiment from the point of view of conflict analysis and future collaboration design.
\citet{zhao2023knowledge} concentrated on detecting causal utterances that lead to the target utterance with a non-neutral emotional tone in conversations, and developed the knowledge-bridged causal interaction network (KBCIN) with commonsense knowledge as bridges.
\citet{chen2024learning} constructed a conversation cognitive model based on intuition theory to explain how each utterance engages and activates information channels in a recursive manner. The authors also synthesize datasets which incorporate implicit causes and complete cause labels.
\citet{zhou-etal-2024-pop} developed the position-oriented prompt-tuning method called POP-CEE to solve the causal emotion entailment task in conversations.


\textbf{Analysis Report Generation.}
\citet{wang2023r2gengpt} explored the potential of LLMs for radiology report generation, and proposed R2GenGPT for generating radiology report.
\citet{ding2024fashionregen} used LLMs to appropriately select and combine multimodal information, integrating text, charts, and images into a coherent, complete, and insightful report. Based on this report, they conducted higher-level understanding and analysis of specific domains.
\citet{wu2024xinyu} developed a LLM-based system Xinyu to assist commentators in generating Chinese commentaries.
\citet{chiang2024badge} focused on badminton report generation, including details such as player names, game scores, and ball types for providing audiences with a comprehensive view, and proposed BADGE based on LLMs.

%
%

\subsection{Skill Enhancement}
\label{tca:se}
After conducting attribution analysis, it is necessary to better utilize these insights to provide targeted training for individuals such as call-center employees or AI agents.
For human participants, some improvement suggestions or suggested wording can be provided, while prompts can be optimized along with some foundational capabilities such as consistency or emotion support for AI agents.
In addition, AI agent can serve as sparring partners to help human enhance skills.

\textbf{Automatic Prompt Optimization.}
Directly Optimizing LLMs for specific tasks may be costly, leading researchers to focus on automatic prompt optimization.
AutoPrompt~\citep{shin-etal-2020-autoprompt} and Prompt-Tuning~\citep{lester-etal-2021-power} firstly attempted to optimize prompts to enhance task performance without training LMs.
\citet{pryzant-etal-2023-automatic} proposed Text Gradient Prompt Optimization (ProTeGi), a general non-parametric algorithm that uses LLM feedback instead of differentiation and LLM editing instead of backpropagation to directly make discrete improvements to prompts.
\citet{NEURIPS2023_c5601d99} introduced a new component called directional stimulus into the prompt, guiding black-box LLMs toward specific desired outputs.
\citet{cheng-etal-2024-black} created an automatic prompt optimizer that rewrites disorganized or unclear human prompts into LLM-preferred prompts for better conveying human intent.

\textbf{Specialized Tuning.} 
1) {\it Conversation Consistency Capability}:
\citet{wang-etal-2024-instruct} improved conversation consistency by separately modeling the speaker roles of agent and user, enabling the agent to adhere to its role consistently.
\citet{zhou-etal-2023-simoap} proposed a two-stage strategy comprising over-sampling and post-evaluation to enhance the consistency of role-based conversation generation.
\citet{shea-yu-2023-building} applied offline reinforcement learning to a role consistency task, demonstrating its ability to enhance role consistency and conversation quality over a system trained solely through imitation learning.

2) {\it Emotion Support Capability}:
\citet{wang-etal-2024-muffin} introduced a novel multifaceted AI feedback mechanism to alleviate the helplessness support framework, aiming to reduce \textit{“unhelpful messages”} in emotional support conversations.
Inspired by the human emotional support generation process of identifying, understanding, and regulating emotions, \citet{zhang-etal-2024-escot} proposed an interpretable emotional support response generation scheme called ESCoT. Based on this, they constructed the first CoT emotional support conversation dataset, ESDCoT.
\citet{zheng-etal-2024-self} utilized LLMs as "Counseling Teacher" to enhance smaller models' emotion support response abilities, significantly reducing the necessity of scaling up model size. 

3) {\it Long-term conversation Capability}:
\citet{Zhong_Guo_Gao_Ye_Wang_2024} enabled LLMs to summon relevant memories, continually evolve through continuous memory updates, comprehend, and adapt to a user’s personality over time by synthesizing information from previous interactions.
\citet{kim2024theanine} proposed a novel timeline-augmented framework, THEANINE, which utilizes a series of memories demonstrating the development and causality of relevant past events to improve long-term conversation capability.
To meet the real-world demands for long-term companionship and personalized interactions with chatbots, \citet{li2024helloagain} introduced a model-agnostic agent framework that simultaneously handles event summary and persona management, enabling the inference of appropriate long-term conversation responses.
%

\textbf{LLM Sparring.}
\citet{li2024leveraging} leveraged the potential of LLMs in clinical medical education, facilitating natural conversations between students and LLM-Simulated Patients, evaluated their conversation, and provided suggestions to enhance students' clinical inquiry skills.
\citet{Lu_2024} found that LLM-simulated students produced logical and believable responses that were aligned with their profiles in multiple-choice question tasks. They exhibit strong consistency with real students' response to the same set of questions. Teachers can use the feedback from LLM-simulated students to improve the quality of the examination.
\citet{xiao-etal-2024-healme} proposed HealMe, a specialized cognitive reframing therapy model that emulates a complete psychotherapeutic procedure, helping clients to better understand their issues, accept new interpretations, and move toward constructive solutions.

\subsection{Conversation Generation}
\label{tca:cg}
Conversation generation focuses on generating concrete conversational content relying on the refined experiences where the core issue is how to make the model follow these experience settings such as the refined scene elements and inject the analyzed insights into the model, steering agents to generate conversations towards specific goals.

\textbf{Setting-Following Based.}
\citet{chen2022weakly} explored few-shot data augmentation for conversation understanding by prompting LLMs and presented a novel approach that iterates on augmentation quality by applying weakly-supervised filters. 
\citet{chen-etal-2023-places} used a small set of expert-written conversations as in-context examples to synthesize a social conversation dataset using prompting.
\citet{zheng-etal-2023-augesc} constructed AugESC, an augmented dataset for the ESC task. They prompted a fine-tuned language model to complete full conversations from available conversation posts of various topics, which are then postprocessed based on heuristics. 
\citet{kim-etal-2023-soda} designed ${CO_3}$ distillation framework and distilled SODA, the first publicly million-scale high-quality social conversation dataset, from the LLM.
\citet{jandaghi-etal-2024-faithful-persona} proposed Generator-Critic architecture framework to create the conversational dataset. The Generator is an LLM prompted to output conversations. The Critic consists of a mixture of expert LLMs that control the quality of the generated conversations.

Various works also used the role-playing abilities of LLMs to generate conversation data. These tasks design conversation elements such as character traits and scenarios, allowing LLM to play specific roles and simulate conversation interactions.
\citet{zhou2023characterglm} designed reasonable attributes and behaviors for AI Characters and adopted a few-shot approach by prompting GPT-4 to generate dialog data.
\citet{lu2024large} introduced Ditto to generate conversation data by self-alignment way, which simulated role-play conversations as a variant of reading comprehension. 
\citet{wang-etal-2024-sotopia} proposed SOTOPIA-$\pi$ method to prompt LLMs generating social interaction (mainly conversations) data based on SOTOPIA \citet{zhousotopia}, then leveraging behavior cloning and self-reinforcement to improve social intelligence ability of LLM.

\textbf{Steering Based.}
\citet{peng-etal-2018-deep} introduced Deep Dyna-Q, a deep RL framework for task-completion conversation policy learning. They integrated a world model into the conversation agent to simulate user responses and experiences. This model is continuously updated with real user data to mimic user behavior better, optimizing the conversation agent with both real and simulated experiences.
\citet{lu2019goal} argued that hindsight experience replay (HER) enables learning from failures, but the vanilla HER is inapplicable to conversation learning due to the implicit goals. They developed two complex HER methods providing different tradeoffs between complexity and performance and enabled HER-based conversation policy learning. 
Strategy learning and goal planning attach great importance in proactive conversation systems. \citet{sanders-etal-2022-towards} conducted a comprehensive evaluation on how LLM-based conversation systems exhibit proactivity, including aspects such as clarification, goal-oriented guidance, and handling of non-cooperative conversations.
\citet{lei2022interacting} focused on proactive conversation policy when users exhibit non-cooperative behavior in interactive environments, balancing between reaching the goal topic and ensuring user satisfaction.
\citet{deng-etal-2023-prompting} introduced the concept of global conversation state and proposed a framework in which conversation agents can understand the trajectory of the ongoing conversation, assess the likelihood of successful outcomes, and evaluate how their own response decisions impact the overall direction of the conversation.

\citet{hong2023zeroshot} explored a new method for adapting LLMs with RL for goal-directed conversation, which collected conversation data through sampling diverse synthetic rollouts of hypothetical in-domain human-human interactions.
\citet{zhu2024emotion} proposed an ESDP, which incorporated user emotions information into conversation policy and selected the optimal action by combining top-k actions with the user emotions. In each turn, the user's emotional information is used as an immediate reward for the current conversation state to solve sparse rewards and the dependency on termination. 
\citet{deng2024plugandplay} introduced PPDPP, a conversation policy planning paradigm that uses a tunable language model plug-in for goal-directed conversation. It offers a training framework for supervised fine-tuning and reinforcement learning from self-play feedback.
\citet{chang-chen-2024-injecting} injected conversation strategies and incorporated chain-of-thought reasoning for training LLMs, where it is equipped with the integrated capabilities
of intent detection, policy selection, and response generation.
\citet{li2024dialogue} presented DAT that adapts language model agents to plan goal-directed conversations. The core idea was to treat each utterance as an action, freeze a pre-trained language model, and train a small planner model that predicts a continuous action vector used for controlled generation in each round.
\citet{zhang2024strength} investigated the limitations of current LLM-based conversation agents in strategic planning, with a key challenge being the agents' awareness of various non-cooperative user behaviors. To address these challenges, they proposed TRIP, which manipulates the experiences of conversation agents, enabling them to recognize the importance of tailoring strategies for individuals.

\subsection{Goal-Directed Optimization}
\label{tca:gdo}
In fact, all conversations are arguably goal-directed~\cite{searle1969speech,austin1975things}, such as helpfulness, harmlessness in open-domain conversation, and completing booking of flight tickets in task-oriented conversation. The goal-directed modeling is gradually gaining attention in open-domain conversation, task-oriented conversation, and conversational recommendation.

\textbf{Open-Domain Conversation.} \citet{tang2019target} studied the conversational goals in open-domain conversation, and expected a conversational system to steer the chat toward a designated target subject by introducing coarse-grained keywords to control the intended content of system responses.
\citet{wu2019proactive} wanted to endow a conversational agent with the ability of proactively leading the conversation, and introduced a new dataset named Konv which consists of a conversation leader and the followers. The leader sequentially changes the discussion topics, following the given conversation goal.
\citet{Ni_Pandelea_Young_Zhou_Cambria_2022} studied how to plan short-term and long-term goal in open-domain conversations, and proposed the hierarchical model to learn goal planning in a hierarchical learning framework.
\citet{yang2022topkg} proposed a global reinforcement learning with planned paths towards the global target.

Recently, goal awareness of open-domain conversation is evoking widespread interest in the era of large language models (LLMs).
\citet{hong2023zeroshot} argued that LLMs can provide useful data for solving goal-directed conversation tasks by simulating sub-optimal but human-like behaviors, instead of effectively solving such tasks out of the box.
\citet{li2024dialogue} adapted language model agents to plan goal-directed conversations by treating each utterance as an action, thereby converting conversations into games and leading to easy adaptation of existing method such as reinforcement learning.
\citet{zhousotopia} introduced SOTOPIA to simulate complex social interactions between LLM agents for achieving goals and evaluate their social intelligence.
\citet{wang-etal-2024-sotopia} further developed SOTOPIA-$\pi$, an interactive learning method, to leverage behavior cloning and self-reinforcement training.

\textbf{Task-Oriented Conversation.} 
\citet{pietquin2006consistent} developed a user modeling technique for realistic simulation of man-machine goal-directed spoken conversations.
\citet{chotimongkol2008learning} argued the purpose of a goal-directed conversation is to fill one or more forms and to ensure that the information is consistent.
~\citet{eshky2014generative} modelled the user behaviour in conversations where user goals are unobserved as part of the conversation.
\citet{Sun_Moon_Crook_Roller_Silvert_Liu_Wang_Liu_Cho_Cardie_2021} integrated task-oriented and open-domain systems to enhance the functional goals (e.g., booking hotels) in task-oriented conversations.
\citet{liu2022go} collected a human-to-human mixed-type task-oriented dialog corpus where an agent provides user-goal-related knowledge to help figure out and achieve specific goals within each session, and then designed a prompt-based continual learning mechanism for model training.

Since the advent of LLMs, 
\citet{ijcai2024p870} used the goal-directed conversation system to infer the current context in situations requiring emergency landings with LLM data augmentations in safety-critical application.
\citet{luo2024duetsim} argued that LLMs were not promising for those responses which effectively guided users towards their goals with intricate constraints and requirements, and employed two LLMs for response generation and verification respectively.
\citet{he-etal-2024-planning} proposed the dual-process conversation planning framework and employed two complementary planning systems inspired by dual-process theory in psychology (thinking—intuitive and analytical) to improve the achievement of goals.

\textbf{Conversational Recommendation.}
\citet{liu2020towards} proposed a conversational recommendation task over multi-type dialogs, with the bots proactively and naturally leading a conversation from a non-recommendation dialog to a recommendation dialog, and developed a human-to-human Chinese dialog dataset DuRecDial.
\citet{hayati2020inspired} created a dataset INSPIRED, which consisted 1,001 human-human dialogs for movie recommendation with measures for successful recommendations.
\citet{zhou2020towards} built a conversational recommendation dataset which leveraged topic threads to enforce natural semantic transitions towards the recommendation scenario.
\citet{bai2021learning} developed goal-oriented knowledge copy network, helping discern those knowledge facts which are highly correlated to the dialog goal and context.
\citet{zhang2021kers} proposed the knowledge-enhanced multi-subgoal driven recommender system to predict a sequence of subgoals which guided the dialog model to select knowledge.
\citet{wang2022follow} aimed to make users accept the goals/targets gradually through conversations, and developed a target-driven conversation planning framework to proactively lead the system to transit between different conversation stages.
\citet{deng2023unified} proposed a unified multi-goal conversational recommender system, to unify {\it goal planning}, {\it topic prediction}, {\it item recommendation}, and {\it response generation} with different formulations into the same sequence-to-sequence paradigm.
\citet{zhang2024goal} introduced a goal interaction graph planning framework to model the correlations and order of goals.

\citet{wang2023rethinking} reviewed the utilization of LLMs for conversational recommendation and showcased the inadequacy of the existing evaluation protocol, and further proposed an interactive evaluation approach via LLM-based user simulators.
\citet{li2024incorporating} aimed to empower LLMs for using external knowledge and goal guidance in conversational recommendation, and decomposed the complex conversational recommendation task into several subtasks through the knowledge retrieval and goal planning.
\citet{huang2024concept} proposed a evaluation protocol on conversational recommendation based on LLMs simulation, which took both system- and user-centric factors into account by conceptualising three key characteristics and further dividing them into six primary abilities. The authors argued that they regarded conversational recommendation as a social issue to consider its major goal of persuading users to accept the recommendation, instead of just a technical problem.

{\it \textbf{Takeaway}}:
An increasing number of research is focusing on goal-awareness ability in conversations, which only models the conversational data to optimize towards the goal based on reinforcement learning, or builds the goal related benchmark and evaluation protocol. 
However, it maybe not promising to lead the model towards a goal solely based on the conversational content without any external information in a black-box manner.
In addition, the goals studied in existing research are relatively shallow and narrow, and lack the universality, most of which aim to achieve a specific goal, such as confirming the harmlessness of the response, or assisting the customer with completing their ticket booking. Differently, CA seeks to analyze out the common, representative and constructive information in conversations for elevating the system's effectiveness in achieving its goals which are generally implicit and grand.
\begin{table*}[htbp]
\renewcommand\arraystretch{0.7}
\tabcolsep=0.14cm
\centering
\tiny
\label{tb-mainresults}
\begin{tabular}{ccccccccccc}
\toprule
{\it \textbf{Section}} &\textbf{Dataset} &\textbf{Parti.} &\textbf{Sce.} &\textbf{Stra.} &\textbf{Go.} &\textbf{Usage} & \textbf{Source} &\textbf{Task} &\textbf{Lang.} &\textbf{Metric} \\ \midrule
\multirow{30}{*}{\makecell[c]{\S \ref{tca:sr}}} 
& {DialogNLI}~\cite{chen-etal-2023-places} &\ding{51} &\ding{55}  &\ding{55} &\ding{55} & \makecell[c]{Train\&Val\\\&Test} & \makecell[c]{Existing Resource} & \makecell[c]{Personal\\ Attribute} & EN & \includegraphics[width=0.015\textwidth]{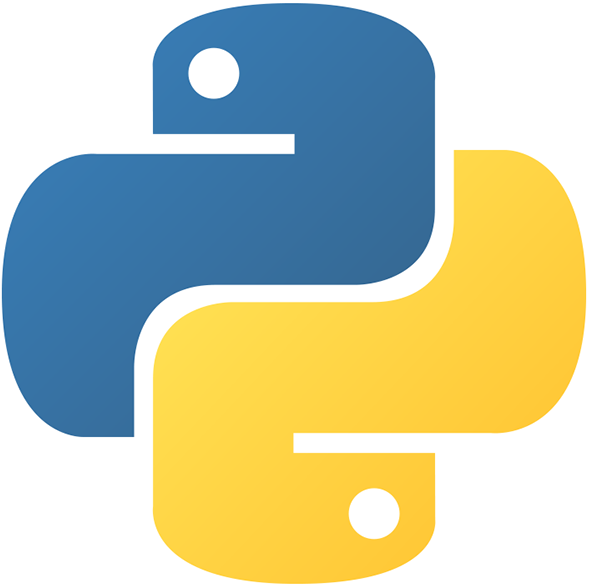} \\
& PersonaChat~\cite{zhang-etal-2018-personalizing} &\ding{51} &\ding{55}  &\ding{55} &\ding{55} & \makecell[c]{Train\&Val\\\&Test} & \makecell[c]{Crowdsourcing} & \makecell[c]{Personal\\ Attribute} & EN & \includegraphics[width=0.015\textwidth]{figure/tubiao-py.png} \includegraphics[width=0.02\textwidth]{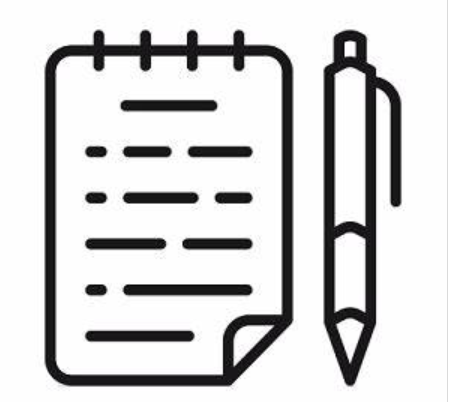}\\
&PersonaExt-PeaCoK~\cite{delucia2024using} &\ding{51} &\ding{55}  &\ding{55} &\ding{55} & \makecell[c]{Train\&Val\\\&Test} & \makecell[c]{Existing Resource} & \makecell[c]{Personal\\ Attribute} & EN & \includegraphics[width=0.015\textwidth]{figure/tubiao-py.png} \includegraphics[width=0.02\textwidth]{figure/tubiao-crit.jpg}\\
&FaceAct$^*$~\cite{sakurai-miyao-2024-evaluating}  &\ding{55} &\ding{51}  &\ding{55} &\ding{55} & \makecell[c]{Test} & \makecell[c]{Existing Resource} & \makecell[c]{Intent} & EN & \includegraphics[width=0.015\textwidth]{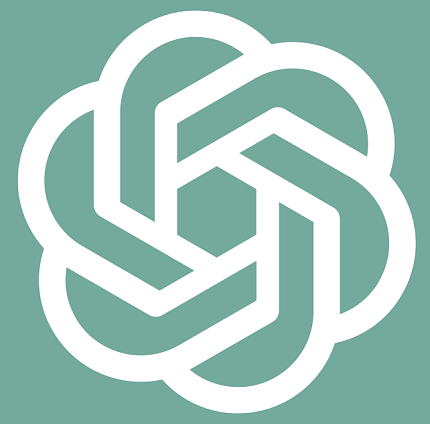} \includegraphics[width=0.02\textwidth]{figure/tubiao-crit.jpg}\\
&Bing Copilot &\ding{55} &\ding{51}  &\ding{55} &\ding{55} & \makecell[c]{Train\&Val\\\&Test} & \makecell[c]{Existing Resource} & \makecell[c]{General} & Multi & \includegraphics[width=0.015\textwidth]{figure/tubiao-py.png}\\
&DialogSum~\cite{chen2021structure} &\ding{55} &\ding{51}  &\ding{55} &\ding{55} & \makecell[c]{Train\&Val\\\&Test} & \makecell[c]{Crowdsourcing} & \makecell[c]{Summarization} & EN & \includegraphics[width=0.015\textwidth]{figure/tubiao-py.png} \includegraphics[width=0.02\textwidth]{figure/tubiao-crit.jpg}\\
&SAMSum~\cite{gliwa2019samsum} &\ding{55} &\ding{51}  &\ding{55} &\ding{55} & \makecell[c]{Train\&Val\\\&Test} & \makecell[c]{Crowdsourcing} & \makecell[c]{Summarization} & EN & \includegraphics[width=0.015\textwidth]{figure/tubiao-py.png} \includegraphics[width=0.02\textwidth]{figure/tubiao-crit.jpg}\\
&CSDS~\cite{lin-etal-2021-csds} &\ding{55} &\ding{51}  &\ding{55} &\ding{55} & \makecell[c]{Train\&Val\\\&Test} & \makecell[c]{Crowdsourcing} & \makecell[c]{Summarization} & ZH & \includegraphics[width=0.015\textwidth]{figure/tubiao-py.png} \includegraphics[width=0.02\textwidth]{figure/tubiao-crit.jpg}\\
&MC~\cite{song2020summarizing} &\ding{55} &\ding{51}  &\ding{55} &\ding{55} & \makecell[c]{Train\&Test} & \makecell[c]{Crowdsourcing} & \makecell[c]{Summarization} & ZH & \includegraphics[width=0.015\textwidth]{figure/tubiao-py.png} \includegraphics[width=0.02\textwidth]{figure/tubiao-crit.jpg}\\
&Clinical-Conversation~\cite{liu-etal-2023-joint} &\ding{55} &\ding{51}  &\ding{55} &\ding{55} & \makecell[c]{Train\&Val\\\&Test} & \makecell[c]{Crowdsourcing} & \makecell[c]{Topic\\ Segmentation} & EN & \includegraphics[width=0.015\textwidth]{figure/tubiao-py.png}\\ 
&SmallDataSet~\cite{canizares2024measuring} &\ding{55} &\ding{51}  &\ding{55} &\ding{55} & \makecell[c]{\&Test} & \makecell[c]{Multi-platform} & \makecell[c]{Topic\\ Clustering} & EN & \includegraphics[width=0.015\textwidth]{figure/tubiao-py.png}\\ 
&AnnoMI~\cite{wu2022anno} &\ding{55} &\ding{55}  &\ding{51} &\ding{55} & \makecell[c]{Test} & \makecell[c]{Crowdsourcing} & \makecell[c]{Strategy} & EN & \includegraphics[width=0.015\textwidth]{figure/tubiaoGPT4.png} \includegraphics[width=0.02\textwidth]{figure/tubiao-crit.jpg}\\
&ABCD~\cite{chen2021action} &\ding{55} &\ding{55}  &\ding{51} &\ding{55} & \makecell[c]{Train\&Val\\\&Test} & \makecell[c]{Crowdsourcing} & \makecell[c]{Strategy} & EN & \includegraphics[width=0.015\textwidth]{figure/tubiao-py.png}\\ \midrule
\multirow{6}{*}{\makecell[c]{\S \ref{tca:ca}}} 
&RECCON-DD~\cite{poria2021recognizing} &\ding{55} &\ding{51}  &\ding{55} &\ding{55} & \makecell[c]{Train\&Val\\\&Test} & \makecell[c]{Crowdsourcing} & \makecell[c]{Emotion Cause} & EN & \includegraphics[width=0.015\textwidth]{figure/tubiao-py.png}\\
&CEDAR~\cite{zhou-etal-2021-probing-commonsense} &\ding{55} &\ding{55}  &\ding{55} &\ding{55} & \makecell[c]{Train\&Val\\\&Test} & \makecell[c]{Crowdsourcing} & \makecell[c]{Commonsense\\ Reasoning} & EN & \includegraphics[width=0.015\textwidth]{figure/tubiao-py.png} \includegraphics[width=0.02\textwidth]{figure/tubiao-crit.jpg}\\
&Synthetic-Simulated~\cite{chen2024learning} &\ding{55} &\ding{51}  &\ding{55} &\ding{55} & \makecell[c]{Train\&Val\\\&Test} & \makecell[c]{Rule\&LLMs} & \makecell[c]{Conversation\\ Reasoning} & EN & \includegraphics[width=0.015\textwidth]{figure/tubiao-py.png} \\
\midrule
\multirow{18}{*}{\makecell[c]{\S \ref{tca:se}}} 
& {\textsc{LIGHT}}~\cite{urbanek2019learning} &\ding{51} &\ding{51}  &\ding{55} &\ding{55} & \makecell[c]{Train\&Val\\\&Test} & \makecell[c]{Crowdsourcing} & \makecell[c]{Conversation Consistency} & EN & \includegraphics[width=0.015\textwidth]{figure/tubiao-py.png} \includegraphics[width=0.015\textwidth]{figure/tubiaoGPT4.png} \\
& TOPDIAL~\cite{wang-etal-2023-target} &\ding{51} &\ding{51}  &\ding{55} &\ding{55} & \makecell[c]{Train\&Val\\\&Test} & \makecell[c]{LLMs} & \makecell[c]{Conversation Consistency} & EN & \includegraphics[width=0.015\textwidth]{figure/tubiao-py.png} \includegraphics[width=0.015\textwidth]{figure/tubiaoGPT4.png} \\
& ESConv~\cite{liu-etal-2021-towards} &\ding{55} &\ding{51}  &\ding{55} &\ding{55} & \makecell[c]{Train\&Val\\\&Test} & \makecell[c]{Crowdsourcing} & \makecell[c]{Emotion Support} & EN & \includegraphics[width=0.015\textwidth]{figure/tubiao-py.png} \includegraphics[width=0.02\textwidth]{figure/tubiao-crit.jpg} \\
&ESD~\cite{zhang-etal-2024-escot} &\ding{55} &\ding{51}  &\ding{51} &\ding{55} & \makecell[c]{Train\&Val\\\&Test} & \makecell[c]{LLMs\&Crowdsourcing} & \makecell[c]{Emotion Support} & EN & \includegraphics[width=0.015\textwidth]{figure/tubiao-py.png} \includegraphics[width=0.02\textwidth]{figure/tubiao-crit.jpg} \\
& {\textsc{LoCoMo}}~\cite{maharana-etal-2024-evaluating} &\ding{51} &\ding{55}  &\ding{55} &\ding{55} & \makecell[c]{Test} & \makecell[c]{Crowdsourcing} & \makecell[c]{Long-Term} & EN & \includegraphics[width=0.015\textwidth]{figure/tubiao-py.png} \\
& MSC~\cite{xu-etal-2022-beyond} &\ding{51} &\ding{51}  &\ding{55} &\ding{55} & \makecell[c]{Train\&Val\\\&Test} & \makecell[c]{Crowdsourcing} & \makecell[c]{Long-Term} & EN & \includegraphics[width=0.015\textwidth]{figure/tubiao-py.png} \includegraphics[width=0.015\textwidth]{figure/tubiaoGPT4.png} \includegraphics[width=0.02\textwidth]{figure/tubiao-crit.jpg}\\
& DuLeMon~\cite{xu2022long} &\ding{51} &\ding{55}  &\ding{55} &\ding{55} & \makecell[c]{Train\&Test} & \makecell[c]{Crowdsourcing} & \makecell[c]{Long-Term} & ZH & \includegraphics[width=0.015\textwidth]{figure/tubiao-py.png} \includegraphics[width=0.02\textwidth]{figure/tubiao-crit.jpg}\\
&CareCall~\cite{bae2022keep} &\ding{55} &\ding{51}  &\ding{55} &\ding{55} & \makecell[c]{Train\&Val\\\&Test} & \makecell[c]{Existing Resource\&\\Crowdsourcing} & \makecell[c]{Long-Term} & Korean & \includegraphics[width=0.015\textwidth]{figure/tubiao-py.png} \includegraphics[width=0.02\textwidth]{figure/tubiao-crit.jpg}\\
& CC~\cite{jang2023conversation} &\ding{55} &\ding{51}  &\ding{55} &\ding{55} & \makecell[c]{Train\&Val\\\&Test} & \makecell[c]{LLMs\&\\Crowdsourcing} & \makecell[c]{Long-Term} & EN & \includegraphics[width=0.015\textwidth]{figure/tubiao-py.png} \includegraphics[width=0.02\textwidth]{figure/tubiao-crit.jpg}\\
\midrule
\multirow{16}{*}{\makecell[c]{\S \ref{tca:cg}}} 
&{WEAKDAP}~\cite{chen2022weakly}  &\ding{55}  &\ding{55} &\ding{55}  &\ding{55} &\makecell[c]{Train} &\makecell[c]{Existing Resource \& LLMs} &Data Augmentation  &EN &\includegraphics[width=0.015\textwidth]{figure/tubiao-py.png} \\

&{PLACES}~\cite{chen-etal-2023-places}  &\ding{51}  &\ding{51} &\ding{55}  &\ding{55} &\makecell[c]{Train} &\makecell[c]{LLMs} &Data Augmentation  &EN &\includegraphics[width=0.02\textwidth]{figure/tubiao-crit.jpg}\\

&{AugESC}~\cite{zheng-etal-2023-augesc}  &\ding{55}  &\ding{55} &\ding{55}  &\ding{55} &\makecell[c]{Train} &\makecell[c]{Existing Resource \& LLMs} &Data Augmentation  &EN &\includegraphics[width=0.02\textwidth]{figure/tubiao-crit.jpg} \includegraphics[width=0.015\textwidth]{figure/tubiao-py.png}\\

&{SODA}~\cite{kim-etal-2023-soda}  &\ding{55}  &\ding{55} &\ding{55}  &\ding{55} &\makecell[c]{Train\&Val\\\&Test} &\makecell[c]{Existing Resource \& LLMs} &Data Augmentation  &EN &\includegraphics[width=0.015\textwidth]{figure/tubiaoGPT4.png}\\

&\makecell[c]{{Synthetic-Persona-Chat}\\~\cite{jandaghi-etal-2024-faithful-persona}}  &\ding{51}  &\ding{55} &\ding{55}  &\ding{55} &\makecell[c]{Train} &\makecell[c]{LLMs} &Data Augmentation  &EN &\includegraphics[width=0.02\textwidth]{figure/tubiao-crit.jpg} \includegraphics[width=0.015\textwidth]{figure/tubiao-py.png}\\

&{CharacterGLM}~\cite{zhou2023characterglm}  &\ding{51}  &\ding{55} &\ding{55}  &\ding{55} &\makecell[c]{Train} &\makecell[c]{Existing Resource \&\\Crowdsourcing \& LLMs} &Role-Playing  &ZH &\includegraphics[width=0.015\textwidth]{figure/tubiaoGPT4.png} \\

&{Ditto}~\cite{lu2024large}  &\ding{51}  &\ding{55} &\ding{55}  &\ding{55} &\makecell[c]{Train\&Test} &\makecell[c]{LLMs} &Role-Playing  &ZH,EN &\includegraphics[width=0.015\textwidth]{figure/tubiaoGPT4.png} \\

&{Sotopia-$\pi$}~\cite{wang-etal-2024-sotopia}  &\ding{51}  &\ding{51} &\ding{55}  &\ding{51} &\makecell[c]{Train} &\makecell[c]{Existing Resource \& LLMs} &Role-Playing  &EN &\includegraphics[width=0.015\textwidth]{figure/tubiaoGPT4.png} \\ 

&{SalesBot 2.0}~\cite{chang-chen-2024-injecting}  &\ding{55}  &\ding{55} &\ding{51}  &\ding{55} &\makecell[c]{Train} &\makecell[c]{Existing Resource \& LLMs} &Strategy Injection  &EN &\includegraphics[width=0.015\textwidth]{figure/tubiaoGPT4.png} \\ \midrule
\multirow{36}{*}{\makecell[c]{\S \ref{tca:gdo}}} & PersonaChat$^*$~\cite{tang2019target} &\ding{55}  &\ding{55} &\ding{55} &\ding{51} &\makecell[c]{Train\&Val\\\&Test}  &\makecell[c]{Existing Resource} & \makecell[c]{Open-Domain} & EN & \includegraphics[width=0.015\textwidth]{figure/tubiao-py.png}\\
&DuConv~\cite{wu2019proactive} &\ding{55}  &\ding{55} &\ding{55} &\ding{51} &\makecell[c]{Train\&Val\\\&Test}  &\makecell[c]{Crowdsourcing} &\makecell[c]{Open-Domain} &ZH &\includegraphics[width=0.015\textwidth]{figure/tubiao-py.png} \\
&OpenDialKG~\cite{moon2019opendialkg} &\ding{55}  &\ding{55} &\ding{55} &\ding{51} &\makecell[c]{Train\&Val\\\&Test}  &\makecell[c]{Crowdsourcing} &\makecell[c]{Open-Domain} &EN &\includegraphics[width=0.015\textwidth]{figure/tubiao-py.png} \includegraphics[width=0.02\textwidth]{figure/tubiao-crit.jpg} \\
&OTTers~\cite{sevegnani2021otters} &\ding{55}  &\ding{55} &\ding{55} &\ding{51} &\makecell[c]{Train\&Val\\\&Test}  &\makecell[c]{Crowdsourcing} &\makecell[c]{Open-Domain} &EN &\includegraphics[width=0.015\textwidth]{figure/tubiao-py.png} \includegraphics[width=0.02\textwidth]{figure/tubiao-crit.jpg} \\
&TGConv~\cite{yang2022topkg} &\ding{55}  &\ding{55} &\ding{55} &\ding{51} &\makecell[c]{Train\&Val\\\&Test}  &\makecell[c]{Existing Resource} &\makecell[c]{Open-Domain} &EN &\includegraphics[width=0.015\textwidth]{figure/tubiao-py.png} \includegraphics[width=0.02\textwidth]{figure/tubiao-crit.jpg} \\
&Sotopia~\cite{zhousotopia} &\ding{51}  &\ding{51} &\ding{55} &\ding{51} &\makecell[c]{Test}  &\makecell[c]{Existing Resource\& LLM} &\makecell[c]{Open-Domain} &EN &\includegraphics[width=0.015\textwidth]{figure/tubiaoGPT4.png} \\
&MultiWOZ~\cite{budzianowski2018multiwoz} &\ding{55}  &\ding{51} &\ding{55} &\ding{51} &\makecell[c]{Train\&Val\\\&Test}  &\makecell[c]{Crowdsourcing} &\makecell[c]{Task-Oriented} &EN &\includegraphics[width=0.015\textwidth]{figure/tubiao-py.png} \\
&SGD~\cite{rastogi2020towards} &\ding{55}  &\ding{51} &\ding{55} &\ding{51} &\makecell[c]{Train\&Val\\\&Test}  &\makecell[c]{Crowdsourcing} &\makecell[c]{Task-Oriented} &EN &\includegraphics[width=0.015\textwidth]{figure/tubiao-py.png} \\
&MultiWOZ 2.1~\cite{eric-etal-2020-multiwoz} &\ding{55}  &\ding{51} &\ding{55} &\ding{51} &\makecell[c]{Train\&Val\\\&Test}  &\makecell[c]{Existing Resource} &\makecell[c]{Task-Oriented} &EN &\includegraphics[width=0.015\textwidth]{figure/tubiao-py.png} \\
&\makecell[c]{ACCENTOR-SGD/\\-MultiWOZ~\cite{Sun_Moon_Crook_Roller_Silvert_Liu_Wang_Liu_Cho_Cardie_2021}} &\ding{55}  &\ding{51} &\ding{55} &\ding{51} &\makecell[c]{Train\&Val\\\&Test}  &\makecell[c]{Existing Resource} &\makecell[c]{Task-Oriented} &EN &\includegraphics[width=0.015\textwidth]{figure/tubiao-py.png} \includegraphics[width=0.02\textwidth]{figure/tubiao-crit.jpg}\\
&DuClarifyDial~\cite{liu2022go} &\ding{55}  &\ding{51} &\ding{55} &\ding{51} &\makecell[c]{Train\&Val\\\&Test}  &\makecell[c]{Crowdsourcing} &\makecell[c]{Task-Oriented} &ZH &\includegraphics[width=0.015\textwidth]{figure/tubiao-py.png} \\
&DuRecDial~\cite{liu2020towards} &\ding{51}  &\ding{55} &\ding{55} &\ding{51} &\makecell[c]{Train\&Val\\\&Test}  &\makecell[c]{Crowdsourcing} &\makecell[c]{Recommendation} &ZH &\includegraphics[width=0.015\textwidth]{figure/tubiao-py.png} \includegraphics[width=0.02\textwidth]{figure/tubiao-crit.jpg}\\
&DuRecDial2.0~\cite{liu2021durecdial} &\ding{51}  &\ding{55} &\ding{55} &\ding{51} &\makecell[c]{Train\&Val\\\&Test}  &\makecell[c]{Crowdsourcing} &\makecell[c]{Recommendation} &ZH,EN &\includegraphics[width=0.015\textwidth]{figure/tubiao-py.png} \includegraphics[width=0.02\textwidth]{figure/tubiao-crit.jpg}\\
&INSPIRED~\cite{hayati2020inspired} &\ding{55}  &\ding{55} &\ding{51} &\ding{51} &\makecell[c]{Train\&Val\\\&Test}  &\makecell[c]{Crowdsourcing} &\makecell[c]{Recommendation} &EN &\includegraphics[width=0.015\textwidth]{figure/tubiao-py.png} \includegraphics[width=0.02\textwidth]{figure/tubiao-crit.jpg}\\
&TG-ReDial~\cite{zhou2020towards} &\ding{51}  &\ding{51} &\ding{55} &\ding{51} &\makecell[c]{Train\&Val\\\&Test}  &\makecell[c]{Crowdsourcing} &\makecell[c]{Recommendation} &ZH &\includegraphics[width=0.015\textwidth]{figure/tubiao-py.png} \includegraphics[width=0.02\textwidth]{figure/tubiao-crit.jpg}\\
&DuConv~\cite{wu2019proactive}  &\ding{55}  &\ding{55} &\ding{55} &\ding{51} &\makecell[c]{Train\&Val\\\&Test}  &\makecell[c]{Crowdsourcing} &\makecell[c]{Recommendation} &ZH &\includegraphics[width=0.015\textwidth]{figure/tubiao-py.png} \includegraphics[width=0.02\textwidth]{figure/tubiao-crit.jpg}\\
\bottomrule
\end{tabular}
\text{Parti., Sce., Stra., and Go. respectively mean {\it Participant}, {\it Scene}, {\it Strategy}, and {\it Goal} elements. \includegraphics[width=0.01\textwidth]{figure/tubiao-py.png}, \includegraphics[width=0.013\textwidth]{figure/tubiao-crit.jpg}, and \includegraphics[width=0.01\textwidth]{figure/tubiaoGPT4.png} respectively represent automatic evaluation such as F1, human, and LLM evaluation.}
\caption{Overview of existing benchmarks and metrics related to CA.}
\end{table*}

\section{Benchmark and Evaluation}
This section assembles the comprehensive datasets pertinent to {\it scene reconstruction}, {\it causality analysis}, {\it skill enhancement}, {\it conversation generation}, and {\it goal-directed optimization}.
Despite the availability of numerous relevant datasets, they only contain ample conversational content and lack detailed conversational scene elements, such as participants (user profile), scenarios, strategies, and goals, which hinders the comprehensive and insightful analyses.

\textbf{Scene Reconstruction.}
Scene elements include participant such as user profile, scenario such as intent, emotion, and strategy. Existing conversation datasets contain some of these elements, but do not include all of them simultaneously. These datasets either contain information about participants~\cite{chen-etal-2023-places,zhang-etal-2018-personalizing,delucia2024using}, scenario elements~\cite{chen2021structure,sakurai-miyao-2024-evaluating} such as intent and summary, or possess some strategy annotations~\cite{chen2021action,wu2022anno}.
Due to the relatively objective nature of these elements, the evaluation metrics are mostly based on hard indicators, such as F1-score, Accuracy, BLEU, and ROUGE-L.
With the development of LLMs, researchers have begun to pay attention to the issue of inconsistency between these indicators and human preferences, and gradually explore LLM evaluators~\cite{wang-etal-2024-large-language-models-fair,zhang2023widerdeeperllmnetworks,zeng2024evaluating,xia-etal-2024-language}.

\textbf{Causality Analysis.}
Currently, causality analysis related conversation datasets are not sufficient. The task definition is actually complex as the causes are more high-level, subjective, and open-ended. To make the task purer and more conducive to manual annotation, existing work tends to simplify the task format. For example, researchers define the emotion cause task, aiming to identify the utterances which stimulate the emotion of the target utterance in conversations~\cite{zhao2023knowledge,zhou-etal-2024-pop}.
Therefore, most of the evaluation criterion are discriminative metrics such as Accuracy and F1-score.

\textbf{Skill Enhancement.}
For AI agents' skill enhancement, existing datasets focus on the improvement of fundamental abilities, such as conversational consistency~\cite{urbanek2019learning,wang-etal-2023-target}, long-term memory~\cite{maharana-etal-2024-evaluating,xu-etal-2022-beyond,xu2022long,bae2022keep,jang2023conversation}, and emotion support~\cite{liu-etal-2021-towards,zhang-etal-2024-escot}. Currently, the benchmarks for automatic prompt optimization and LLM sparring in conversations are still scare. The evaluation metrics are some common criteria such as F1-score, Accuracy, BLEU, and ROUGE-L, and gradually introducing LLM evaluators.

\textbf{Conversation Generation.}
In order to evaluate and enhance the LLMs' ability to follow the role settings, some research efforts have been made in dataset construction.
These datasets~\cite{zhou2023characterglm,chen-etal-2023-places,jandaghi-etal-2024-faithful-persona,lu2024large} are built by the following ways: 1) recruiting crowd-sourcing workers for playing the setting, such as role or character, 2) prompting LLMs such as GPT-4 to generate synthetic data, 3) extracting conversations for specific characters from literary sources like novels.
Since the quality of conversations, especially whether they adhere to the relevant settings, is highly subjective, most evaluations are conducted by pre-defining several evaluation dimensions or aspects (such as consistency, engagement) and then using powerful LLMs such as GPT-4 or human to evaluate. They mostly use scores (such as 1 to 5) or pairwise battles between models to quantify the performance.
In addition, for steering conversation generation, most datasets are sourced from existing proactive conversation datasets~\cite{deng2024plugandplay, chang-chen-2024-injecting}. Besides the quality of conversations, goal completion is also considered.

\textbf{Goal-Directed Optimization.}
In open-domain conversation, task-oriented conversation, and conversational recommendation, a substantial amount of relevant datasets have been accumulated with the task development.
These datasets have either been derived from existing datasets~\cite{tang2019target,yang2022topkg,Sun_Moon_Crook_Roller_Silvert_Liu_Wang_Liu_Cho_Cardie_2021} or specifically constructed through crowdsourcing~\cite{wu2019proactive,moon2019opendialkg,sevegnani2021otters,budzianowski2018multiwoz,rastogi2020towards,liu2022go} such as human-to-human written conversations to meet the requirements of the task, where the goals are shallow and are a series of keywords, and the goals are considered met if these keywords are mentioned in the conversation.
In addition, the datasets are predominantly in English, and the evaluation metrics primarily consist of hard measures, such as Accuracy, F1-score, Success Rate, BLEU, ROUGE-L, hits@K, PPL, and soft ones such as human and LLM evaluations. These metrics are tailor-designed and exploited for evaluating the achievement of the goals. For instance, success rate refers to the fraction of conversations that successfully accomplish the user's task~\cite{luo2024duetsim}.

\section{Discussions}
The above sections have provided an overview of the current landscape in the research field related to CA. Following this, the discussion will summarize the evolving trends in current research. Additionally, this section will outline potential avenues for future research, aiming to address gaps and explore new horizons in the field.

\subsection{Research Trends}
Current work is not systematic and the corresponding techniques are scattered and fragmented, but there is an overarching trend moving towards the following advancements:
\begin{itemize}
    \item \textbf{Task Formulation:} there is a shift from \textcolor{gray}{\it requiring human intervention to decompose tasks into atomic ones with strict input-output formats} towards \textcolor{darkgray}{\it \textbf{more natural language interactions with loose ones}}. This implies that tasks are becoming more flexible and necessitate the enhanced instruction-following ability of models. For example, a lot of work~\cite{louvan2020recent} in the past required a clear distinction between discriminative and generative tasks, with rigidly defined input and output formats, making it difficult to handle open-ended problems and resulting in poor scalability. However, LLMs adopt a unified instruction-following and generative paradigm, which enhances the flexibility and openness of task modeling~\cite{wang-etal-2022-extracting,zhang2023safeconv}.
    \item \textbf{Task Complexity:} we observe an evolution from \textcolor{gray}{\it surface-level character reassembly} to \textcolor{darkgray}{\it \textbf{deeper and implicit semantic restructuring}}. This requires models to better understand the subtleties and implied meanings within conversations. In earlier studies, researchers focus on ASR error correction~\cite{mani2020towards} or coreference resolution~\cite{zhang2021did,kim2021learn} in conversations, which requires a relatively superficial understanding to achieve promising performance. However, real-world conversations often contain many expressions that carry implications beyond the literal words, such as irony, which presents a significant challenge to comprehending the conversations and receives more attention in recent years~\cite{li2023diplomat,ruis2023thegoldilocks,yue2024large}.
    \item \textbf{Task Modeling:} there is a transition from \textcolor{gray}{\it third-person point of view} to \textcolor{darkgray}{\it \textbf{first-person interactive simulation modeling}}. This necessitates greater generalization, transferability, and deductive reasoning capabilities within the models. In previous studies, the model directly takes the conversations as input to obtain the representations or generate chain-of-thought style rationales for specific tasks~\cite{lin2022other,feng2023schema,zhang-etal-2024-escot} under the third-person perspective. With the development of LLMs, some research efforts try to situate the model in a specific task context for task solution under the first-person perception paradigm, such as by simulating the process of data generation and outputting the relevant analysis results by the way~\cite{liu2023one,wang-etal-2024-reasoning,niu-etal-2024-enhancing}, which have a stronger sense of immersion.
\end{itemize}

\subsection{Future Directions}
This subsection will explore emerging frontiers in CA research, with the aim of inspiring future investigations in this field.

\textbf{LLM Conversation Simulator.}
With the development of LLMs which gradually gain powerful instruction-following abilities, researchers start to explore simulation modeling of LLMs.
The simulation is mainly divided into two parts: 1) conversational content simulator where the model serves as a conversational AI agent~\cite{li-etal-2022-controllable}, and 2) conversational scene simulator where the model acts as a conversational analyzer~\cite{wang-etal-2024-reasoning}.
The conversational content simulator generates conversation data according to relevant settings, which have attracted a huge amount of interest in the era of LLMs, such as role-playing, conversation data augmentation.
The conversational scene simulator mines useful conversational elements such as user profile, emotion, strategy, summarization by simulating conversations instead of providing chain-of-thought style rationales. The simulation process integrates the model into specific conversational contexts, thereby enhancing its situational awareness and in-depth analyses. Therefore, how to construct the conversational scene simulator for conversation analysis is worth exploring.

\textbf{Fine-Grained Conversation Benchmark.}
Although tasks involving conversations have been extensively studied, the accumulated datasets are only applicable to one or a few specific atomic tasks~\cite{delucia2024using} and pay more attention to conversation generation~\cite{zhousotopia}.
Currently, there is a lack of conversation analysis datasets that contain comprehensive and fine-grained conversational scene elements (participant profile, topic, emotion, strategy, etc.), which hinders the evaluation and modeling of conversation analysis, such as the ability to understand conversational scenes and perform the causality analysis.
Therefore, the research communities urgently need high-quality and comprehensive conversation analysis benchmark.

\textbf{Long-Context Conversation Modeling.}
Conversation analysis takes the conversational content as input and also faces the long-context challenge, such as the loss of key information, confusion between the parties (attributing one participant's viewpoint to another) in conversations.
Although the long-context modeling has been extensively studied in documents~\cite{wang2024leavedocumentbehindbenchmarking,he-etal-2024-never,chen-etal-2024-long}, LLMs encounter the significantly different challenges in long-context conversational content from the documents, such as contextual response dependency and inconsistency as the participants may change their minds.
Therefore, it is necessary to conduct specialized research on long-context modeling in conversations.

\textbf{In-Depth Conversation Attribution.}
Most research related to conversation analysis tends to uncover the relatively shallow elements, such as emotion~\cite{lin-etal-2024-interpretable}, intent~\cite{song-etal-2022-enhancing}, and summarization~\cite{tian-etal-2024-dialogue}, which leads to the significant gap between the analysis results and business application. The business personnel are not clear on how to utilize these superficial results for empowering the business innovation.
What they really need are the underlying reasons behind these shallow results, along with some constructive suggestions.
For example, in customer service conversations, while models can detect the user's angry emotion, the causes of this frustration and how to prevent these occurrences in subsequent services remain under-explored.
However, it is extremely challenging to perform these in-depth attribution explorations in the era of small language models due to their limited modeling ability, while the breakthrough in the large language models brings a ray of hope for in-depth analyses thanks to their powerful instruction-following ability and abundant world knowledge. Building an ecosystem for evaluating and modeling in-depth attribution is vital, such as introducing real-world tasks with corresponding benchmarks.

\textbf{Goal-Directed Conversation Optimization \& Evaluation.}
Although goal-directed conversation modeling has been extensively explored in recent years, most of these goals are generally simple and restricted to a certain scenario, such as mentioning specific keywords~\cite{tang2019target} or persuading users to accept recommendations~\cite{wang2022follow,li2024incorporating} within a conversation.
In fact, real-world business goals are often very complex and abstract within multiple conversations, such as improving the customer's experience and increasing user utilization, which raises higher requirements for the inductive and deductive abilities of models. How to bridge the gap between these goals and the analysis results from models is worth further exploring.
In addition, these goals are usually abstract and high-level, so traditional metrics such as F1-score, BLEU cannot accurately measure the achievement of goals, and LLM evaluators also face significant challenges due to the abstractness. Therefore, the ability to evaluate goal achievement for LLM evaluators is of great importance and requires additional investigation.

\textbf{Cross-Session Conversation KV Cache.} The reuse of key-value cache has always been one of the key technologies for reducing inference costs~\cite{NEURIPS2023_a452a7c6,chen-etal-2024-nacl}. From inter-sentence eviction to the recent cross-layer reuse, it can effectively reduce the cost of a single inference. 
However, the number of users using LLM-based conversation systems is gradually increasing, and the conversation content between different users and LLMs is not completely isolated; there are many similar topics. 
Efficiently reusing cross-session key-value cache and effectively storing the conversation history cache of different types of users will be key to improving the efficiency of conversation systems and reducing costs.

\textbf{Conversation Security.}
The emergence of LLMs has made the conversation a  predominant medium of interaction, both among people and between humans and machines. The security of conversations should receive increasing attention.
As shown in Figure~\ref{ca_diagram}, CA is the analysis process in continuous and iterative manners, centered around specific goals such as improving user satisfaction or discovering violations.
As CA systems continuously analyze out critical information and give insights for reducing violations, the evolution of non-compliant behavior to resist regulation will be ongoing. Therefore, the constant evolution of the CA system in such an adversarial context is worth exploring, such as in the self-play manner~\cite{zhang2024survey}.
In addition, the extreme imbalance between non-compliant and normal samples in conversations poses serious challenges for modeling.
Fortunately, LLMs make it possible to simulate the non-compliant data, enhancing the modeling capability for non-compliant samples.
The exploration of how to jailbreak LLMs~\cite{NEURIPS2023_fd661313,zhou2024alignmentjailbreakworkexplain} to generate these positive examples that are beneficial for modeling has a wide range of needs.
\section{Conclusion}
This paper presents the first endeavor to outline conversation analysis (CA) and perform a relevant literature review from a technical viewpoint. CA is framed by four principal steps centered around the goal achievement: (1) scene reconstruction, (2) causality analysis, (3) skill enhancement, and (4) conversation generation, which reveals the enormous application value of CA in the era of rapid development of language UI under the backdrop of LLMs. We have also gathered benchmarks related to CA and provided some discussions on development trends and future directions, hoping to kick-start and accelerate the rapid development and business application of CA.
\section*{Limitations}
We provide a comprehensive review on recent techniques related to CA, and make a first attempt to define the CA task from the viewpoint of technique, including four critical procedures. We have also re-organized the existing data resources and pointed out the potential value of CA in the context of business goal achievement.
Nevertheless, we only make the high-level descriptions on existing work related to scene reconstruction, causality analysis, skill enhancement, conversation generation, and goal-directed optimization. In the next version, we will give detailed analyses in each procedure and build an arena to conduct in-depth comparisons of various techniques.

\bibliography{ref}
\end{document}